\documentclass[11pt,a4paper]{article}
\usepackage[hyperref]{emnlp2020}
\usepackage{times}
\usepackage{latexsym}

\usepackage{microtype}

\usepackage{booktabs}
\usepackage{multirow}
\usepackage{graphicx} 
\usepackage{subcaption}
\usepackage{comment}

\usepackage{amsmath}
\usepackage{amsfonts}
\usepackage{amssymb}

\usepackage{algorithm}
\usepackage{algorithmic}

\aclfinalcopy 
\def\aclpaperid{76} 

\title{
ALICE: Active Learning with Contrastive Natural Language Explanations 
}

\author{Weixin Liang \\
  Stanford University \\
  \texttt{wxliang@stanford.edu} \\\And
  James Zou$^{1}$ \\
  Stanford University \\
  \texttt{jamesz@stanford.edu} \\\And
  Zhou Yu$^{1}$ \\
  University of California, Davis \\
  \texttt{joyu@ucdavis.edu} \\
  }

\date{}

\newcommand{\hightlightmodelname}[1]{{{{
\underline{A}ctive \underline{L}earning 
w\underline{i}th 
\underline{C}ontrastive 
\underline{E}xplanations
}}}{#1}}
\newcommand{\modelabbrevname}[1]{{{{ALICE}}}{#1}}

\begin{document}
\maketitle

\begin{abstract}


Training a supervised neural network classifier typically requires many annotated training samples. Collecting and annotating a large number of data points are costly 
and sometimes even infeasible. 
Traditional annotation process uses a low-bandwidth human-machine communication interface: 
classification labels, each of which only provides a few bits of information. 
We propose \hightlightmodelname{} 
 (\modelabbrevname ), 
an expert-in-the-loop training framework 
that utilizes  
contrastive natural language explanations 
to improve data efficiency in learning. 
\modelabbrevname{}
learns to first use active learning to select the most informative  pairs of label classes to elicit contrastive natural language explanations from experts. Then it extracts knowledge from these explanations using a semantic parser. Finally, it incorporates the extracted knowledge through dynamically changing the learning model's structure. We applied \modelabbrevname{} in two visual recognition tasks, bird species classification and social relationship classification. 
We found by incorporating contrastive explanations, our models outperform baseline models that are trained with 40-100\% more training data. 
We found that adding $1$ explanation 
leads to similar performance gain 
as adding 13-30 labeled training data points.


\end{abstract}
\footnotetext[1]{Co-supervised project.}

\section{Introduction}
The de-facto supervised neural network training paradigm requires a large dataset with annotations. 
It is time-consuming, difficult and sometimes even infeasible to collect a large number of data-points 
due to task nature. A typical example task is medical diagnosis. 
In addition, annotating datasets also is costly, especially in domains where experts are difficult to recruit. In a traditional annotation process, the  human-machine communication bandwidth is narrow. Each label provides $\log C$ bits per sample for a $C$-class classification problem. 
However, humans don't solely rely on such low bandwidth communication to learn. They instead learn through natural language communication, which grounds on abstract concepts and knowledge. Psychologists and philosophers have long posited 
natural language explanations 
as central, organizing elements to human learning and reasoning~\cite{chin2017contrastive,lombrozo2006structure,smith2003learning}. Following this intuition, we explore methods to incorporate natural language explanations in learning paradigms to improve learning algorithm's data efficiency.

\begin{figure}[t]
\centering
\includegraphics[width=0.40\textwidth]
{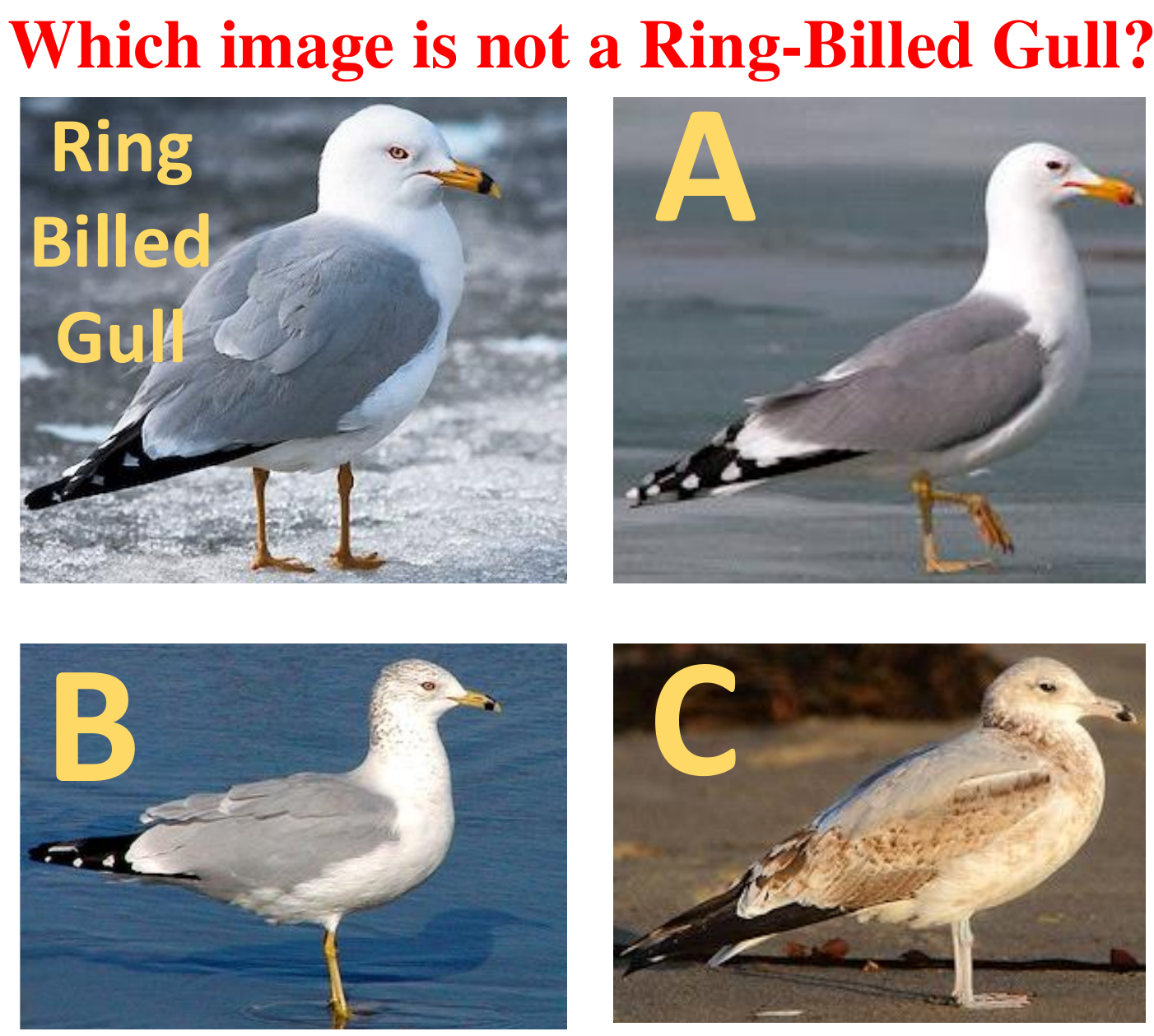} 
\vspace{-2mm}
\caption{
\small
An example task that would benefit from learning with natural language explanation. 
The top-left corner shows an example image of a ring-billed gull. In the other three images (A), (B), (C), which one is not a ring-billed gull but a California gull? 
Given the natural language explanation  
\textit{``Ring-billed gull has a bill with a black ring near the tip while California gull has a red spot near the tip of lower mandible''}, it would be easier to find that (A) is the correct choice. 
}
\vspace{-7mm}
\label{fig:motivating}
\end{figure}

Let's take a bird species classification task as an example to illustrate the advantage of learning with natural language explanation. Figure~\ref{fig:motivating} shows several bird images. 
Based on visual dissimilarity, 
many people mistakenly thought Image C is not a ring-billed gull as it has a different colored coat compared to the example. However, ring-billed gulls change their coat color from light yellow to grey after the first winter. So color is not the deciding factor to distinguish California gull and ring-billed gull.
If we receive abstract knowledge from human experts through  a natural language format, such as \textit{``Ring-billed gull has a bill with a black ring near the tip while California gull has a red spot near the tip of lower mandible''} and incorporate it in the model, then the model will
discover that Image A is a California gull instead of a ring-billed gull based on 
its bill.

Previous work has shown that 
incorporating natural language explanation into the classification training loop 
is effective in various settings~\cite{LearningWithLatentLanguage,mu2020shaping}. 
However, 
previous work neglects the fact that 
there is usually a limited time budget 
to interact with domain experts (e.g., medical experts, biologists)~\cite{MOSS,DBLP:conf/acl/LiangZY20} 
and high-quality natural language explanations are expensive, by nature. 
Therefore, 
we focus on eliciting fewer but more informative explanations to reduce expert involvement.

We propose \hightlightmodelname{} 
 (\modelabbrevname ), 
an expert-in-the-loop training framework 
that utilizes 
contrastive natural language explanations 
to improve data efficiency in learning. 
Although we focus on image classification in this paper, 
our expert-in-the-loop training 
framework could be generalized to other classification tasks. 
\modelabbrevname{}
learns to first use active learning to select the most informative query pair to elicit contrastive natural language explanations from experts. Then it extracts knowledge from these explanations using a semantic parser. Finally, it incorporates the extracted knowledge through dynamically updating the learning model's structure. 
Our experiments on bird species classification and social relationship classification show that our method that incorporates natural language explanations has better data efficiency compared to methods that increase training sample volume.

\begin{figure*}[tb]
\centering
\includegraphics[width=1.0\textwidth]
{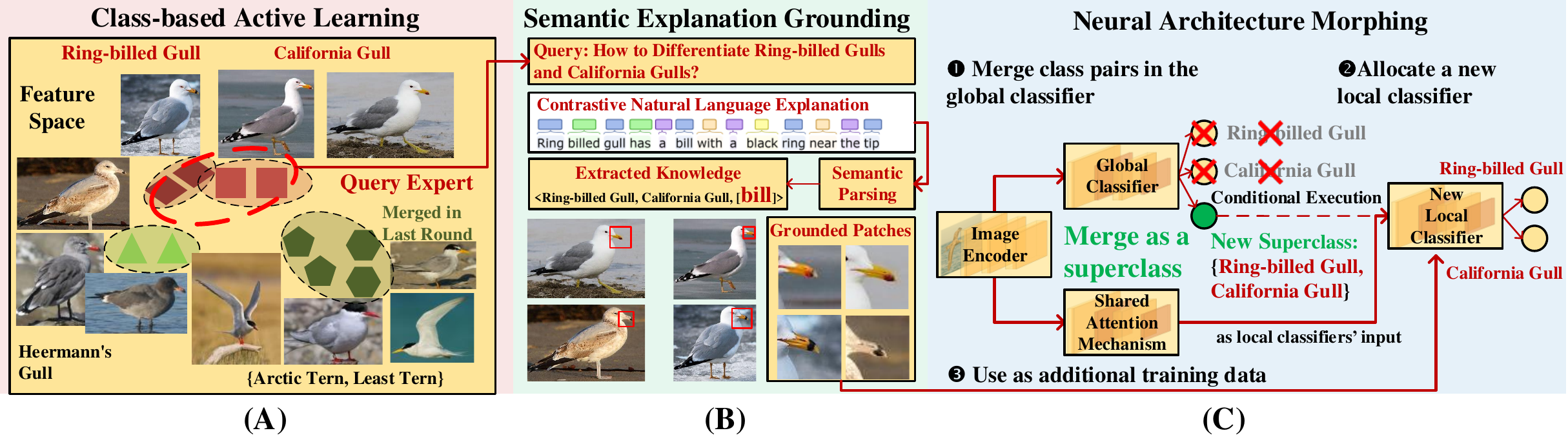} 
\vspace{-6mm}
\caption{
\small
\modelabbrevname's three-step workflow 
for each round. 
\textbf{(A) Class-based Active Learning}: 
\modelabbrevname{} first
projects
each class's training data into a shared feature space. 
Then \modelabbrevname{} selects $b$ most confusing class pairs 
to query domain experts for explanations. 
\textbf{(B) Semantic Explanation Grounding}: 
\modelabbrevname{} then extracts knowledge from 
$b$ contrastive natural language explanations 
by semantic parsing. 
\modelabbrevname{} 
grounds 
the extracted knowledge 
on the training data of $b$ class pairs 
by cropping 
the corresponding semantic segments. 
\textbf{(C) Neural Architecture Morphing}: 
\modelabbrevname{} finally
allocates $b$ new local classifiers and 
merges $b$ class pairs in the global classifier. 
The cropped image patches are used 
as additional training data 
for a newly added local classifier to emphasize these patches' importance. 
The model is re-trained after each round. 
}
\vspace{-5mm}
\label{fig:main}
\end{figure*}

\section{Related Work}

\paragraph{Learning with Natural Language Explanation}
Psychologists and philosophers have long posited 
natural language explanations 
as central organizing elements to human learning and reasoning~\cite{chin2017contrastive}. 
Several attempts have been made to incorporate natural language explanations into supervised classification tasks. 
\citet{LearningWithLatentLanguage,mu2020shaping} 
adopt a multi-task setting by learning classification and captioning simultaneously. 
\citet{ExpBERT,DBLP:conf/cvpr/HeP17} encode natural language explanations as additional features to 
assist classification. 
Orthogonal to their approaches, 
we focus on eliciting fewer but more informative explanations to reduce expert involvement with class-based active learning. 
Another line of research collects heuristic rules as explanations (e.g., `honey month' for predicting SPOUSE relationship) 
to automatically label unlabeled data~\cite{DBLP:conf/emnlp/SrivastavaLM17,renxiang,NaturalLanguageExplanations}. 
Different from their settings, 
we assume no additional training data-points. 
In addition, we leverage natural language explanations by extracting knowledge 
and incorporate the knowledge into classifiers. 
Distantly related to our work, 
\citet{DBLP:conf/eccv/HendricksARDSD16} propose to generate explanations for image classifiers but they do not explore improving the classifiers with the explanations.

\paragraph{Active Learning} 
The key hypothesis of active learning is that, 
if the learning algorithm is allowed to choose the data from which it learns, 
it will perform better than randomly selecting training samples ~\cite{settles2009active}. 
Existing work in active learning focuses primarily on 
exploring sampling methods to select additional data-points to label from a pool of unlabeled data~\cite{DBLP:conf/iclr/SenerS18,settles2011theories,settles2009active}. 
~\citet{groupbased} propose group-based active learning where the annotator could label a group of data points each time rather than one data point. 
However, they still rely on classification labels 
as the interface for 
human-machine communication. 
Instead, we focus on 
incorporating natural language explanations into the classification training framework. 
Contrastive learning has previously been shown to substantially improve unsupervised learning \cite{abid2018exploring}, feature learning \cite{zou2015crowdsourcing}, and learning probabilistic models \cite{zou2013contrastive}. However, it has not been applied to the setting of active learning with explanations as we explore here.

\paragraph{Hierarchical Visual Recognition}
Categorical hierarchy is inherent 
in visual recognition~\cite{biederman1987recognition,ijcv}. 
\citet{DBLP:conf/mm/XiaoZYPZ14} propose to expand the model 
based on category hierarchy for incremental learning. 
\citet{HDCNN} decompose classification task into  a coarse category classification and a fine category classification. 
Different from previous work, 
we focus on incorporating contrastive natural language explanations into the model hierarchy to achieve better data efficiency.

\section{Problem Formulation}

\paragraph{Contrastive Natural Language Explanations} 
\label{subsec:Explanations}
Existing research in social science and cognitive science 
\cite{miller2019explanation,DBLP:conf/fat/MittelstadtRW19} 
suggests contrastive explanations are more effective in human learning than descriptive explanations. 
Therefore, we choose contrastive natural language explanations to benefit our learners. An example contrastive explanation is like  ``Why P rather than Q?'', in which P is the target event and Q is a counterfactual contrast case that did not occur~\cite{lipton1990contrastive}. 
In the example in Figure~\ref{fig:motivating}, 
if we ask the expert to differentiate between Ring-billed gull against California gull, 
the expert would output the following natural language explanation: 
\textit{``Ring-billed gull has a bill with a black ring near the tip while California gull has a red spot near tip of lower mandible''}. 
Our explanations are class-based 
and are not specifically associated with any particular images.

\paragraph{Problem Setup} 
\label{subsec:Problem Setup}

We are interested in a $C$ class classification problem defined
over an input space $X$ 
and a label space $Y = \{1, ... , C\}$. 
Initially, 
the training set 
    $ D_{train} = \{(x_i, y_i)\}^{N_{train}}_1 $ 
    is small, since our setting is restricted to be low resource. 
We also assume that there is 
a limited budget to ask domain experts to provide explanations during training. 
Specifically, 
we consider $k$ rounds of interactions with domain experts 
and each round has a query budget $b$. 
For each query, 
we need to specify two classes $y^p, y^q$ for domain experts to compare. 
Domain experts would return 
a contrastive natural language explanation $e$. 
Each explanation $e$ would guide us to focus on the most discriminating semantic segments to differentiate between $y^p$ and $y^q$.  
In this paper, a
\textit{semantic segment} refers to a semantic segment of an object (e.g., ``bill'' in bird species classification) or 
a semantic object (e.g., ``soccer'' in social relationship classification).

To make our framework more general, 
we start from a standard image classification neural architecture.  
We formulate our initial model as $M(\phi, g_{pool}, f) = f(g_{pool}(\phi(x)))$: 
Here $\phi$ is an image encoder that maps each input image $x$ to 
    an activation map $\phi(x) \in \mathbb{R}^{H \times W \times d}$.  
$g_{pool}$ is a global pooling layer 
    $ g_{pool}(\phi(x)) \in \mathbb{R}^{d_{pool}}$.
$f$ is a fully connected layer 
    that performs flat $C$ way classification. 
This formulation covers most of the off-the-shelf pre-trained image classifiers. 

\section{\modelabbrevname: \hightlightmodelname{} }

\subsection{Overview} 

\modelabbrevname{} is an expert-in-the-loop training framework 
that utilizes 
contrastive natural language explanations 
to improve data efficiency in learning. 
\modelabbrevname{} performs multiple rounds of interaction with domain experts and 
dynamically updates the learning model's structure 
during each round. 
Figure~\ref{fig:main} describes 
\modelabbrevname's three-step workflow 
for each round:  
\textbf{(A) Class-based Active Learning}: 
\modelabbrevname{} first 
projects
each class's training data into a shared feature space. 
Then \modelabbrevname{} selects $b$ most confusing class pairs 
to query domain experts for explanations. 
\textbf{(B) Semantic Explanation Grounding}: 
\modelabbrevname{} then extracts knowledge from 
$b$ contrastive natural language explanations 
by semantic parsing. 
\modelabbrevname{} 
grounds 
the extracted knowledge 
on the training data of $b$ class pairs 
by cropping 
the corresponding semantic segments. 
\textbf{(C) Neural Architecture Morphing}: 
\modelabbrevname{} finally
allocates $b$ new local classifiers and 
merges $b$ class pairs in the global classifier. 
The cropped image patches are used 
as additional training data 
for a newly added local classifier to emphasize these patches' importance. 
The model is re-trained after each round.

\subsection{Class-based Active Learning} 
\label{subsec:Class-based Active Learning}

\modelabbrevname{} 
optimizes towards 
requesting the most informative explanations 
to reduce expert involvement. 
Since each explanation provides knowledge to distinguish a class pair, 
we aim to identify 
the class pairs that confuse the model most 
and the explanations on these class pairs 
would intuitively help the model a lot. 
\modelabbrevname{} identifies confusing class pairs by 
first projecting 
each class's training data into a shared feature space $g_{pool}(\phi(x))$. 
As shown in Figure~\ref{fig:main}~(A), 
if the training data of two classes 
are close in the feature space, 
it is usually hard for the model to distinguish them 
and thus it would be helpful to solicit an explanation on this class pair. 
Based on this intuition, 
we first define the distance between two classes and then select the class pairs with the lowest distance. 
We first profile each class $j$ 
by fitting a multivariate Gaussian distribution $\mathcal{N}_j(\mu_j, \, \Sigma_j)$ 
on class $j$'s training sample features. 
We define the distance between class $j$ and class $k$ as 
the Jensen–Shannon Divergence (JSD) 
between $\mathcal{N}_j$ and $\mathcal{N}_k$. 
\begin{equation*}
\mathbb D_{\mathsf J} (\mathcal{N}_j, \mathcal{N}_k) \triangleq 
\frac{1}{2} 
\mathbb D_{\mathsf{KL}} (\mathcal{N}_j || (\mathcal{N}_{jk}) )
+ 
\frac{1}{2} 
\mathbb D_{\mathsf{KL}} (\mathcal{N}_k  || \mathcal{N}_{jk} ) 
\end{equation*}
where $\mathcal{N}_{jk} = \frac{1}{2}(\mathcal{N}_j + \mathcal{N}_k) $ and 
$\mathbb D_{\mathsf{KL}} (\mathcal{N}_j || \mathcal{N}_k )$ is the Kullback-Liebler (KL) divergence: 
\begin{align*}
    \mathbb D_{\mathsf{KL}} 
    (\mathcal{N}_j || \mathcal{N}_k) = &  
    \frac{1}{2}
    \Big( 
        tr(\Sigma_k^{-1} \Sigma_j - I) + \log(\frac{|\Sigma_k|}{|\Sigma_j|}) \\
        & + 
        (\mu_j - \mu_k)^T \Sigma_k^{-1} (\mu_j - \mu_k) 
    \Big)
\end{align*}
After calculating the distance between all possible class pairs,  
we select the $b$ class pairs with the lowest JSD distance to query domain experts.

\subsection{Semantic Explanation Grounding}
\label{subsec:Semantic Explanation Grounding} 

After identifying $b$ class pairs that the model is most confused about, 
we send $b$ query to domain experts. 
We ask the expert the following question for each query, ``How would you differentiate class $P$ and class $Q$?''. 
Since we want the expert to provide general class-level knowledge, 
each query only contains text information, and no visual examples are provided to the experts. 
We obtain $b$ contrastive natural language explanations after the query. 
Next, we parse the natural language explanations into 
machine-understandable form.

\begin{table}[htb]
\small
\centering
\begin{tabular}{p{7cm}}
\cmidrule[\heavyrulewidth]{1-1}
\textbf{Query Expert:} 
``How to differentiate \textcolor{orange}{\textit{Ring-billed Gulls}} and 
\textcolor{blue}{\textit{California Gulls}}?'' \\
\cmidrule{1-1} 
\textbf{Parse Expert Explanation:} ``\textcolor{orange}{\textit{Ring-billed Gull}} has 
a bill with a black ring near the tip while \textcolor{blue}{\textit{California Gull}} has a red spot near the tip of lower mandible. ''\\
\textbf{Extracted Knowledge:} Pay attention to \textcolor{red}{\textbf{[Bill]}} when classifying \textcolor{orange}{\textit{Ring-billed Gull}} v.s. \textcolor{blue}{\textit{California Gulls}}
\\
\cmidrule{1-1} 
\textbf{Ground Extracted Knowlwdge:} Crop \textcolor{red}{\textbf{[Bill]}} in every training image of \textcolor{orange}{\textit{Ring-billed Gulls}} and 
\textcolor{blue}{\textit{California Gulls}} 
\begin{minipage}{.435\textwidth}
  \includegraphics[width=\linewidth]{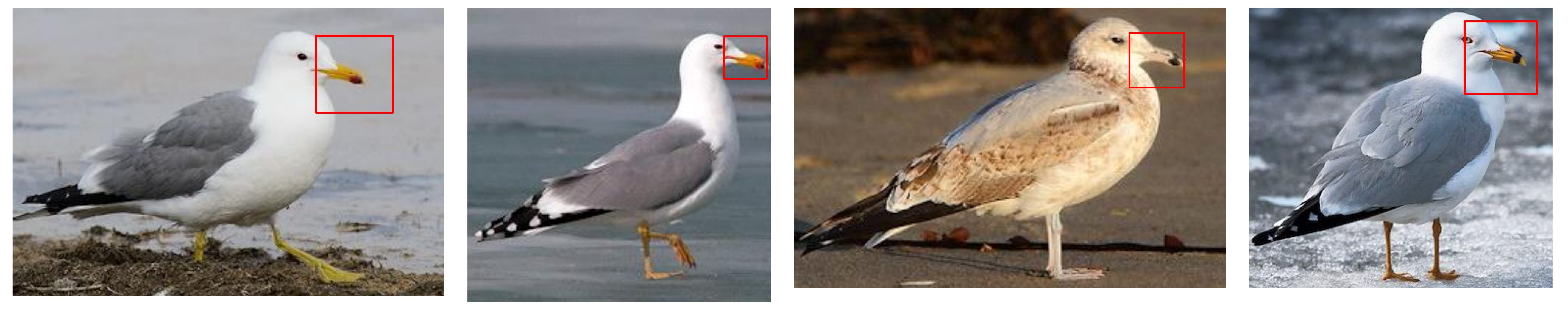} 
\end{minipage}
\\
\cmidrule[\heavyrulewidth]{1-1}
\end{tabular}
\vspace{-2mm}
\caption{
\small 
Semantic Explanation Grounding Workflow
}
\label{table:semanticExplanationGrounding}
\vspace{-2mm}
\end{table}

We choose a simple rule-based semantic parser 
for simplicity, 
following~\citet{NaturalLanguageExplanations}. 
The simple rule-based semantic parser 
can be used without any additional training 
and requires minimum effort to develop. 
Formally, the parser uses a set of rules in the form $\alpha \rightarrow \beta$, which means that $\alpha$ can be replaced by the token(s) in $\beta$. 
Our rules focus primarily on identifying the discriminating semantic segments~($\S$~\ref{subsec:Explanations}) mentioned in the explanations (e.g., ``bill'' for differentiating between ring-billed gull and California gull). 
We also allow the parser to skip unexpected tokens 
so that the parser could always succeed in generating a valid output. 

Since each explanation $e$ provides class-level knowledge to distinguish class $y^p, y^q$, 
we need to propagate the knowledge 
to all the training data-points in class $y^p, y^q$ 
so that the learning model could incorporate the knowledge later during training. 
We denote the semantic segments mentioned in an explanation $e$ 
as $S = \{s_1, s_2, ..., \}$. 
For each training data-point of class $y^p, y^q$, 
we apply off-the-shelf semantic segment localization models 
to crop out 
the image patch(es) of the semantic segment(s) mentioned $S = \{s_1, s_2, ..., \}$ (Figure~\ref{fig:main}~(B)). 
The number of patches cropped from each image equals the number of mentioned semantic segments (i.e., $|S|$). 
We then resize the image patches to full resolution. 
The intuition behind our crop-and-resize approach comes from the popular image crop data augmentation: it augments the training data with ``sampling of various sized patches of the image whose size is distributed evenly between 8\% and 100\% of the image area''~\cite{DBLP:conf/cvpr/SzegedyLJSRAEVR15}. This data augmentation technique is widely-adopted and is supported by common deep learning frameworks like PyTorch\footnote{\texttt{torchvision.transforms.RandomResizedCrop}}.

\modelabbrevname{} does not need the localization model during testing (More details in $\S$~\ref{subsec:morphing}). 
The off-the-shelf semantic segment localization models 
could be the pre-trained localization models on 
various large-scale datasets 
like Visual Genome~\cite{DBLP:journals/ijcv/KrishnaZGJHKCKL17} and PASCAL-Part~\cite{DBLP:conf/cvpr/ChenMLFUY14}. 
If there is no available off-the-shelf localization model, 
we could recruit non-expert annotators 
to annotate the location of the semantic segments 
given that our training set $D_{train}$ is small.

\subsection{Neural Architecture Morphing}
\label{subsec:morphing}

\paragraph{Overview} 
\modelabbrevname{} incorporates 
contrastive natural language explanations 
through dynamically updating the learning model's structure. 
The high-level idea is to 
allocate a number of local classifiers 
to help the origin model 
guided by the explanations. 
Specifically, 
for each explanation $e$ that provides knowledge to 
distinguish two classes $y_p, y_q$, 
we allocate a local classifier that is dedicated 
to the binary classification between $y_p, y_q$. 
We incorporate the extracted knowledge from explanation $e$ 
to the local classifier 
so that the local classifier learns to focus 
on the discriminating semantic segments 
pointed out by the domain experts. 
We first discuss the case where 
all local classifiers perform binary classification 
and then discuss how to extend them 
to support general m-ary classification.

\paragraph{Progressive Architecture Update} 
The initial flat $C$-way classification architecture could be viewed as a composition of 
an image encoder $\phi$ and a global classifier $f \circ g_{pool}$. 
We discuss how the local classifiers are progressively added to assist the global classifier. 
As shown in Figure~\ref{fig:main} (C), 
we first merge $b$ class pairs 
into $b$ super-classes 
in the global classifier.  
For example, 
in the first round, 
the global classifier would change 
from $C$-way to $(C-2b+b)$-way. 
We then allocate 
$b$ new local classifiers, 
each for performing binary classification for one class pair. 
Each local classifier is only called when  the global classifier 
predicts its super-class as the most confident. 
We delay more complex conditional execution schemes as future work. 
We also note that the conditional execution schemes have potential for reducing computation runtime~\cite{FrugalML,deepstore}. 
During training, we fine-tune the image encoder $\phi$ and reset the global classifier after each round since it is only a linear layer. 

\paragraph{Knowledge Grounded Training} 
The global classifier is trained on $D_{train}$, 
with labels adjusted according to the class pair merging. 
For a local classifier 
corresponding to the class pair $y^p, y^q$, 
its training data consists of two parts. 
One part of the training data is the training data-points of classes $y^p, y^q$ in $D_{train}$. 
The other part is the resized image patches of class $y^p, y^q$ 
obtained in semantic explanation grounding~($\S$~\ref{subsec:Semantic Explanation Grounding}). 
We use the resized image patches as additional training data to to emphasize these patches' importance. 
Take the local classifier distinguishing ring-billed gull and California gull as an example (Figure~\ref{fig:main}~(B, C)). 
This local classifier is trained on 
the training images of ring-billed gull and California gull, 
as well as 
the bills' patches of each training image of ring-billed gull and California gull. 
During testing, we only feed the whole image into the model. 

\paragraph{Supporting m-ary local classifier} 
So far we have assumed that 
the local classifier is always a binary classifier. 
An implicit assumption is that 
the $b$ class pairs have no overlap. 
We could support overlapping class pairs as follows. 
If some class pairs have overlap (e.g., class pair $(P,Q)$, class pair $(P,T)$, class pair $(T, U)$), 
we only allocate one local classifier for them (e.g., a $4$-ary local classifier for class $(P,Q,T,U)$). 
We also merge all the relevant classes 
in the global classifier into only one super-class (e.g., super-class $\{P,Q,T,U\}$). 
The local classifier is trained on the 
union of 
the overlapping class pairs' training data including patches.

\begin{figure}[tb]
\centering
\includegraphics[width=0.45\textwidth]
{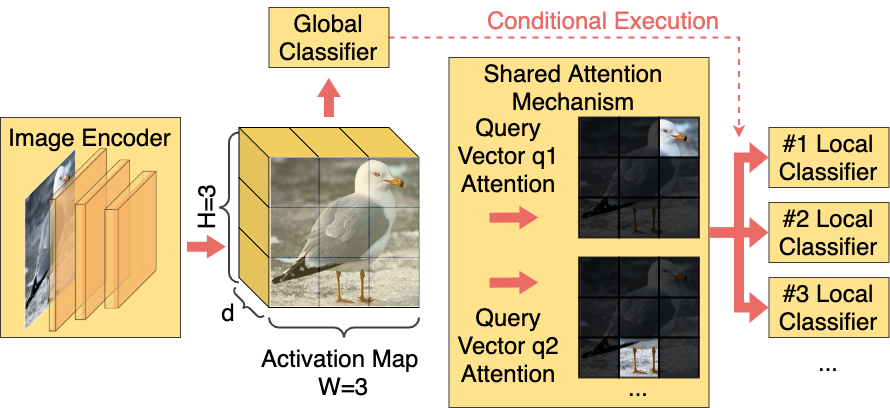} 
\vspace{-3mm}
\caption{
\small 
Local classifiers with shared attention mechanism 
}
\vspace{-5mm}
\label{fig:neuralArch}
\end{figure}

\paragraph{Local Classifier Design} 
Our framework is agnostic 
to the design choice of the local classifiers. 
Any design could be plugged into \modelabbrevname. 
We provide a default design as follows. 
Ideally, 
each local classifier should learn  
which semantic segments to focus and 
how to detect them. 
Since different local classifier 
might need to detect the same semantic segments (e.g., bill), 
the knowledge of detecting semantic segments could be shared among all local classifiers. 
Therefore, 
we introduce a shared attention mechanism, 
which is parameterized using 
$M$ learnable latent attention queries $q_1, q_2, ..., q_M \in \mathbb{R}^{d}$ 
that represent $M$ different latent semantic segments. 
To keep our design general, 
we do \emph{not} bind each latent attention queries to any concrete semantic segments 
(e.g., we do not assign binding like $q_1$ to ``bill'') 
and these queries are trained 
in a weakly-supervised manner.  
Following~\citet{DBLP:conf/iccv/LinRM15,DBLP:journals/corr/abs-1901-09891}, 
we view the activation map 
$\phi(x) \in \mathbb{R}^{H \times W \times d}$ of each image $x$ 
as $H \times W$ attention keys $k_1, ..., k_{H \times W} \in \mathbb{R}^{d}$. We compute the attention by: 
\vspace{-4mm}
\begin{equation*}
Q={ 
\left[ \begin{array}{c}
q_1^T \\
... \\ 
q_M^T 
\end{array} 
\right ]}, 
K=V={ 
\left[ \begin{array}{c}
k_1^T \\
... \\ 
k_{H \times W}^T 
\end{array} 
\right ]}
\end{equation*}
\vspace{-4mm}
\begin{equation*}
   A = \mathrm{Attention}(Q, K, V) = \mathrm{softmax}(\frac{QK^T}{\sqrt{d}})V 
\end{equation*}
Where $Q \in \mathbb{R}^{M \times d}$, $K=V \in \mathbb{R}^{(H\times W) \times d}$. 
Each row in the attention output matrix $A \in \mathbb{R}^{M \times d}$ 
is the attention output for each attention query $q_i$, 
which is a descriptor of the $i^{th}$ latent semantic segments. 
After the shared attention mechanism, 
each local classifier applies a private fully-connected layer on $\mathrm{flattened}(A)$ to make predictions. 
Each local classifier could ignore irrelevant semantic segments 
by simply setting the corresponding weights 
in its fully-connected layer to zero.

\paragraph{Implementation} 
Our image encoder
$\phi$ could be any off-the-shelf visual backbone model and we use Inception v3~\cite{DBLP:conf/cvpr/SzegedyVISW16}. 
We implement our semantic parser on top of the Python-based SippyCup~\cite{liang2015bringing} 
following previous work~\citet{NaturalLanguageExplanations}. 
Our framework could support applications in other languages by changing a semantic parser for corresponding languages. 
We provide more details in Appendix. 

\section{Bird Species Classification Task}
\label{sec:Bird Species Classification Task}

\begin{figure}[tb]
\centering
\includegraphics[width=0.48\textwidth]
{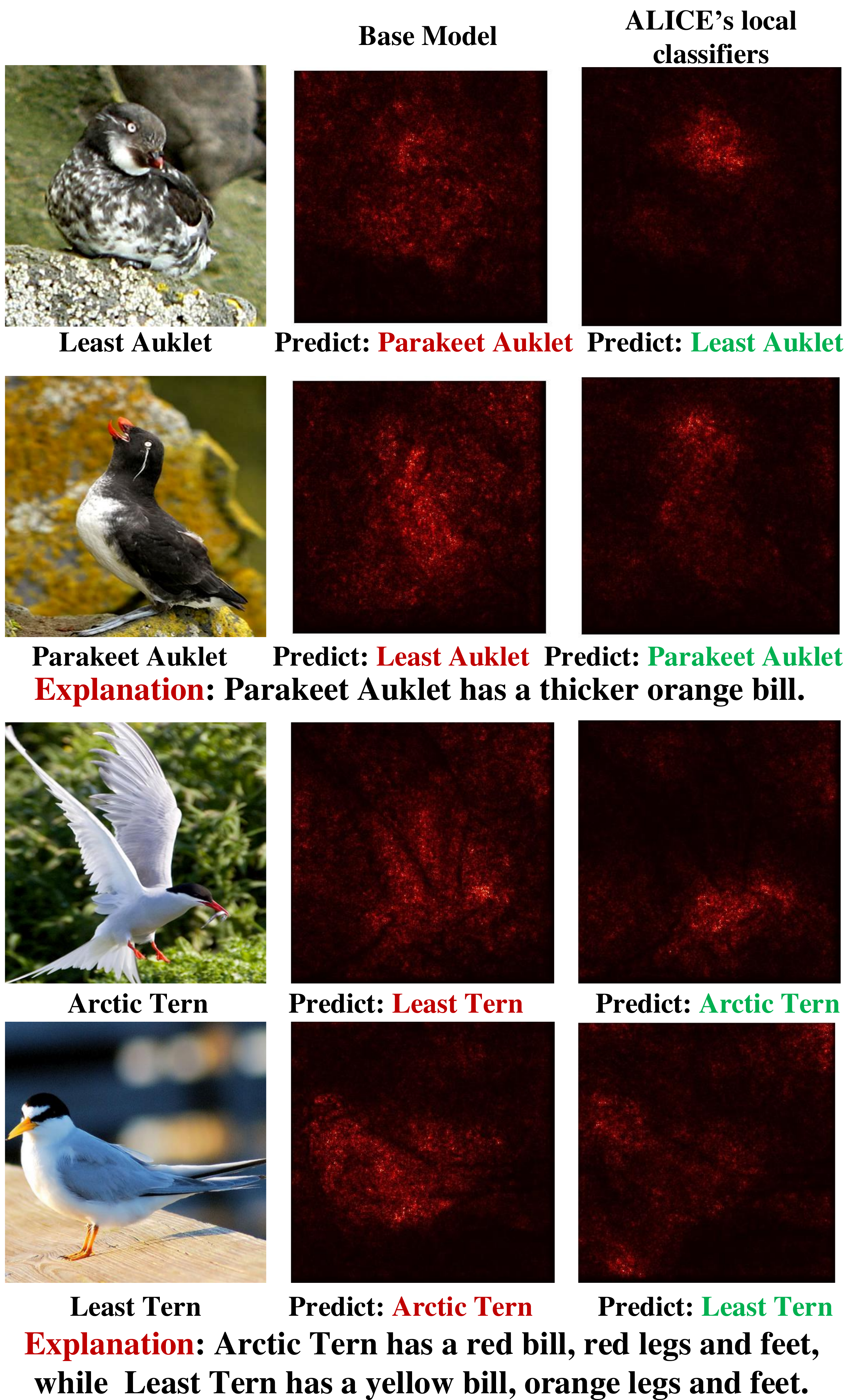} 
\vspace{-4mm}
\caption{
\small 
Saliency maps visualization. 
Guided by expert explanations, 
\modelabbrevname{} 
learn to focus on the discriminating 
semantic segments and make the correct prediction. 
}
\vspace{-4mm}
\label{fig:visualize}
\end{figure}

\paragraph{Dataset}
We use the CUB-200-2011 
dataset~\cite{WahCUB_200_2011}, which contains $11,788$ images for $200$ species of North American birds. 
We randomly sample $25$ bird species due to limited access to expert query budget. 
Following \citet{DBLP:conf/cvpr/VedantamB0PC17},  
We make sure that each sampled species has one or more confusing species 
from the same \textit{subfamilia} 
so that they are challenging to classify. 
In addition, each image in the CUB data-set is also annotated with 
the locations of 
$15$ semantic segments (e.g., ``bill'', ``eye'').  
We use these location annotations to crop training image patches based on the explanations. 
We do not use any location annotation during testing. 
More details are provided in the Appendix, including the list of $25$ sampled species. 
We experiment with a low-resource setting 
with only $15$ images per bird species.

We employ an amateur bird watcher as the domain expert 
since we do not expect 
general MTurk workers to have enough domain expertise. 
To further ensure the annotation quality,  
our domain expert checks  
the professional birding field guide 
~\footnote{
\url{https://identify.whatbird.com/}} 
before writing each explanation. 
We ask the expert, ``How would you differentiate bird species $P$ and bird species $Q$?''. 
In total, we collect $67$ 
contrastive natural language explanations (avg. length $18.45$ words). 
We collect the explanations in an on-demand manner because our class-based active learning is empirically insensitive 
to the change of random seeds and hyper-parameters. 
Our semantic parser identifies $2.36$ semantic segments per explanation on average. 
In each experiment, 
we conduct $k=4$ rounds of expert queries, 
with a query budget $b=3$ for each round.

\paragraph{Discussion on CUB Description Dataset} 
The CUB description dataset collects descriptions of visual appearance for each image rather than explanations of why the bird in the image belongs to a certain class~\cite{DBLP:journals/corr/ReedASL16,DBLP:conf/eccv/HendricksARDSD16}. For example, an image with a Ring-billed gull has the description: \textit{``This is a white bird with a grey wing and orange eyes and beak.''} However, this description also fits perfectly with a California gull (Figure~\ref{fig:motivating}). So the crowd-sourced descriptions in the CUB description dataset is not ideal to support classification. We collected expert explanations: \textit{``Ring-billed gull has a bill with a black ring near the tip while California gull has a red spot near the tip of lower mandible.''} to improve classification data efficiency. 
In addition, we also conducted experiments to incorporate CUB descriptions (5 sentences per image), but we did not find improved performance in our setting.

\begin{figure}[t]
\centering
\includegraphics[width=0.40\textwidth]{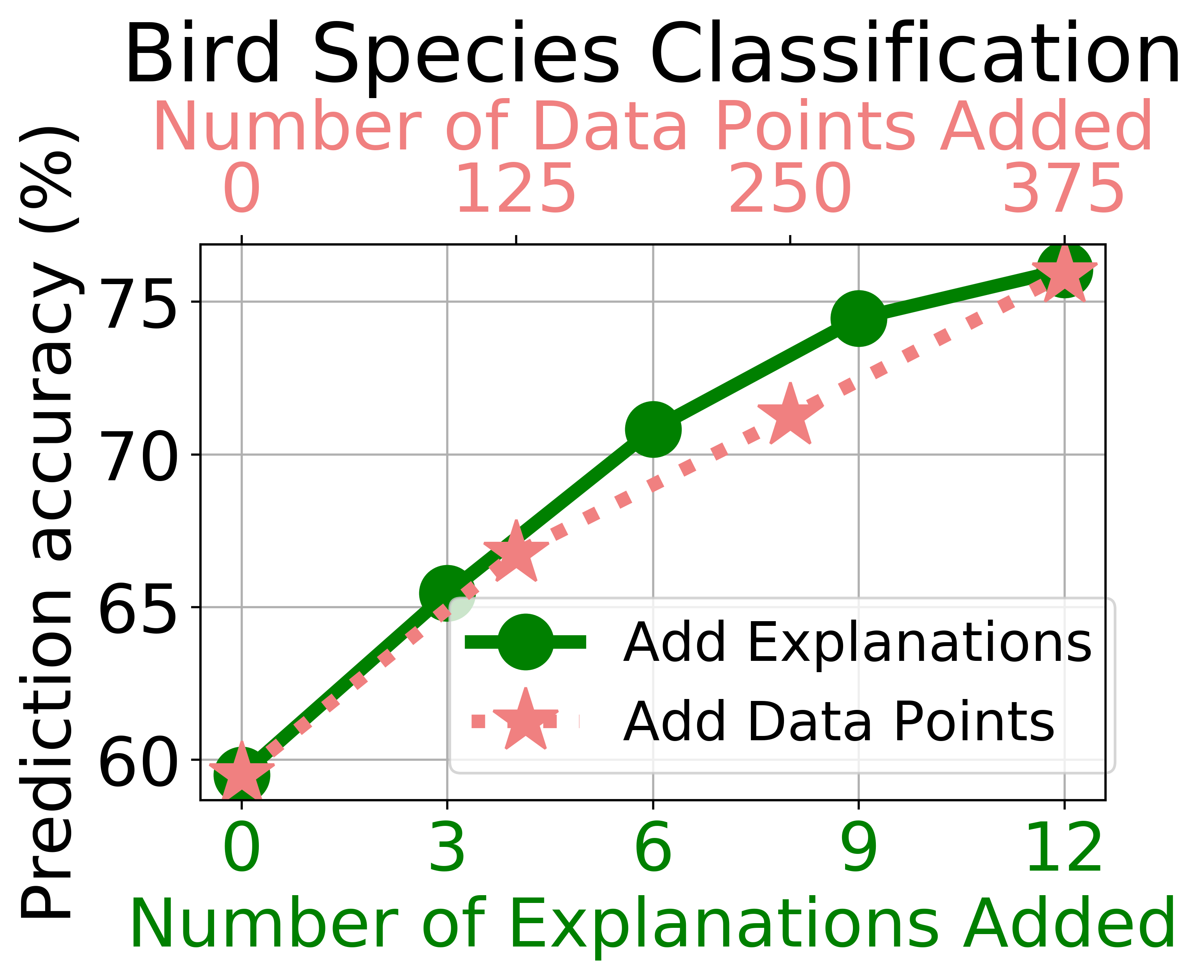}
\caption{
\small
Comparing the performance gain of adding contrastive natural language explanations and adding training data points on bird species' prediction accuracy. 
Empirically, 
adding $1$ explanation 
leads to similar performance gain 
as adding $30$ labeled training data points. 
}
\label{fig:dataEfficiency} 
\vspace{-5mm}
\end{figure}

\paragraph{Model Ablations and Metrics} 
We compare \modelabbrevname{} to its several ablations (Table~\ref{tab:mainresult}) and evaluate the performance 
on the test set. 
We report classification accuracy 
on species as well as subfamilia. 
For subfamilia accuracy, 
a prediction is counted as correct as long as the predicted species' subfamilia is the same as 
the labeled species' subfamilia. 
(1) \textit{Base(Inception v3)} 
fine-tunes the pre-trained Inception v3 
to perform a flat-$25$ way classification. 
(2) \textit{\modelabbrevname{} w/o Grounding} 
copies the final neural architecture from \modelabbrevname{} 
but does not have access to 
the discriminating semantic segments ($\S$~\ref{subsec:Semantic Explanation Grounding}). 
(3) \textit{\modelabbrevname{} w/o Hierarchy} 
has the same neural architecture as (1) 
but has access to the discriminating semantic segments. 
(4) \textit{\modelabbrevname{} w/ Random Grounding} 
has the semantic segments that are randomly sampled. 
(5) \textit{\modelabbrevname{} w/ Random Pairs} 
replaces class-based active learning with randomly selected class pairs. 
The randomly selected class pairs are used to query experts and 
change the learning model's neural architecture. 
(9-12) \textit{\modelabbrevname{}~$i^{th}$~round} 
shows \modelabbrevname's performance after the $i^{th}$ round of expert queries. 
(6-8) \textit{RandomSampling~+~$x\%$~extra~data} 
augments (1) with $x\%$ extra training data points.

\begin{table}[t]
\small
\begin{center}
\tabcolsep=0.08cm
\begin{tabular}{lrll}
\cmidrule[\heavyrulewidth]{1-4}
\multirow{2}{*}{\bf No.} 
& \multirow{2}{*}{\bf Model} 
&\multicolumn{2}{l}{\bf Accuracy (\%)} 
\\
\cmidrule{3-4}
 & & \textit{species} & \textit{subfamilia} \\
\cmidrule[\heavyrulewidth]{1-4}
(1) & Base(Inception v3) & 59.51 & 86.50 \\
\cmidrule{1-4} 
(2) & \modelabbrevname{} w/o Grounding & 66.47 & 87.95 \\ 
(3) & \modelabbrevname{} w/o Hierarchy & 59.22 & 86.94 \\ 
(4) & \modelabbrevname{} w/ Random Ground & 64.44 & 87.52 \\ 
(5) & \modelabbrevname{} w/ Random Pairs & 42.67 & 75.33 \\ 
\cmidrule{1-4} 
(6) & RandomSampling~+~$33\%$~extra~data & 66.76 & 88.39 \\
(7) & RandomSampling~+~$66\%$~extra~data & 71.26 & 91.00 \\
(8) & RandomSampling~+~$100\%$~extra~data & 75.91 & 91.58 \\
\cmidrule{1-4} 
(9) & \modelabbrevname{} ($1^{st}$~round) & 65.46 & 86.07 \\  
(10) & \modelabbrevname{} ($2^{nd}$~round) & 70.83 & 89.84 \\  
(11) & \modelabbrevname{} ($3^{rd}$~round) & 74.46 & 91.00 \\  
(12) & \bf \modelabbrevname{} ($4^{th}$~round) &\bf 76.05 & \bf 91.87  \\  
\cmidrule[\heavyrulewidth]{1-4}
\end{tabular}
\vspace{-4mm}
\caption{
\small
Test accuracy comparison among variants of \modelabbrevname{} 
on the bird species classification task. 
}
\label{tab:mainresult}
\end{center}
\vspace{-7mm}
\end{table}

\paragraph{Results} 
Our first takeaway is that 
incorporating 
contrastive natural language explanations 
is more data-efficient 
than adding extra training data points. 
Figure~\ref{fig:dataEfficiency} visualizes 
the performance gain of adding explanations and adding data points. ((6-12) in Table~\ref{tab:mainresult}). 
As shown in Figure~\ref{fig:dataEfficiency}, 
adding $1$ explanation leads to the same amount of performance gain of adding $30$ labeled data points. 
For example, 
adding $12$ explanations (\modelabbrevname{} ($4^{th}$~round), $76.05\%$) 
achieves comparable performance gain 
of adding $375$ training images 
(RandomSampling~+~$100\%$~extra~data, $75.91\%$). 
We note that writing one explanation for an expert 
is typically faster than labeling 15-30 examples. 
As an estimate, 
\citet{renxiang,NaturalLanguageExplanations,DBLP:conf/emnlp/ZaidanE08} 
perform user study and 
find that collecting natural language explanations is only 
twice as costly as collecting labels for their tasks. 
Our experiment shows that adding $1$ explanation leads to similar performance gain as adding $30$ labeled training data points, yielding a $6\times$ speedup.

Our second takeaway is that 
both the grounding of explanations' semantics and 
the hierarchical neural architecture 
improves classification performance a lot. 
Removing the grounded training image patches degrades 
\modelabbrevname's performance (\modelabbrevname{} w/o Grounding, $66.47\%$). 
Substituting the discriminating semantic segments' image patches 
with other semantic segments' patches leads to worse performance 
(\modelabbrevname{} w/ Random Grounding, $64.44\%$). 
The hierarchical neural architecture is also important. 
As shown in Table~\ref{tab:mainresult}, 
a baseline model augmented with hierarchical classification (\modelabbrevname{} w/o Grounding, $66.74\%$) outperforms 
the flat $C$ way classification (Base(Inception v3), $59.51\%$). 
Similarly, removing the hierarchical neural architecture from \modelabbrevname{} drops the performance a lot (\modelabbrevname{} w/o Hierarchy, $59.22\%$ v.s. \modelabbrevname{} ($4^{th}$~round), $76.05\%$). 
\modelabbrevname{} morphs the neural architecture 
based on class-based active learning ($\S$~\ref{subsec:Class-based Active Learning}). 
If we replace class-based active learning with a random selection of class pairs, 
\modelabbrevname{}
learns a bad model structure that leads to reduced performance (\modelabbrevname{} w/ Random Pairs, $42.67\%$).

\begin{table}[tb]
\small
\begin{center}
\begin{tabular}{lrll}
\cmidrule[\heavyrulewidth]{1-4}
\multirow{2}{*}{\bf No.} 
& \multirow{2}{*}{\bf Method} 
&\multicolumn{2}{l}{\bf Species Acc (\%)} 
\\
\cmidrule{3-4}
 & & \textit{+33\% data} & \textit{+66\% data} \\
\cmidrule[\heavyrulewidth]{1-4}
(1) & RandomSampling & 66.76 & 71.26 \\
(2) & CoreSet & 68.06 & 73.09 \\
(3) & LeastConfidence & 67.34 & 71.94 \\
(4) & MarginSampling & 66.04 & 70.36 \\
(5) & EntropySampling & 66.91 & 72.52 \\
(6) & BALDdropout & 66.33 & 71.65 \\
\cmidrule[\heavyrulewidth]{1-4}
\end{tabular}
\vspace{-2mm}
\caption{
\small
Instance-based active learning baselines on the bird species classification task. 
We note that \modelabbrevname{} ($4^{th}$~round, Acc $76.05\%$) in Table~\ref{tab:mainresult}~(12) 
outperforms all instance-based active learning baselines with $66\%$ extra training data 
(Acc $70.36\%$-$73.09\%$).
}
\label{tab:activelearningresult}
\end{center}
\vspace{-7mm}
\end{table}

\paragraph{Additional Experiments} 
Table~\ref{tab:activelearningresult} shows our experiments with several common instance-based active learning baselines.  
We show the test accuracy of adding $33\%$ extra training data (i.e., $125$ extra data points) and adding $66\%$ extra training data (i.e., $250$ extra data points) 
using the instance-based active learning baselines. 
In this case, 
we observe that \modelabbrevname{} with 12 explanations (Accuracy $76.05\%$, Table~\ref{tab:mainresult}~(12)) outperforms all instance-based active learning baselines with 250 extra data points(Accuracy $70.36\%$-$73.09\%$, Table~\ref{tab:activelearningresult}). 
We delay the combination of instance-based active learning and our class-based active learning as future work. 
To testify whether \modelabbrevname{} could work robustly with smaller amount of training data, 
we present an experiment on CUB starting with as few as $5$ images per species. 
\modelabbrevname{} with $12$ explanations ($k=4$, $b=3$) improves the accuracy of the base model from $49.76\%$ to $62.80\%$, outperforming the base model with $15$ images per class (Accuracy $59.51\%$, Table~\ref{tab:mainresult}).

\paragraph{Visualization} 
We show how the explanations help the learning model as shown in Figure~\ref{fig:visualize}. 
We visualize the saliency maps~\cite{saliencymap} corresponding to the correct class on four example images. 
As shown in Figure~\ref{fig:visualize}, 
the base model does not know which semantic segments to focus 
and makes wrong predictions. 
In contrast, 
\modelabbrevname's local classifiers 
obtain knowledge from the expert explanations 
and successfully learns to focus on the discriminating 
semantic segments to make the correct predictions.

\section{Social Relationship Classification Task}

\begin{table}[t]
\small
\begin{center}
\tabcolsep=0.11cm
\begin{tabular}{lrll}
\cmidrule[\heavyrulewidth]{1-4}
\multirow{2}{*}{\bf No.} 
& \multirow{2}{*}{\bf Model} 
&\multicolumn{2}{l}{\bf Accuracy (\%)} 
\\
\cmidrule{3-4}
 & & \textit{relation} & \textit{domain} \\
\cmidrule[\heavyrulewidth]{1-4}
(1) & Base(Inception v3) & 33.67 & 45.39 \\
\cmidrule{1-4} 
(2) & \modelabbrevname{} w/ Random Ground & 27.20 & 42.52 \\
(3) & \modelabbrevname{} w/ Random Pairs & 22.94 & 35.29 \\
\cmidrule{1-4} 
(4) & RandomSampling~+~$20\%$~extra~data & 34.91 & 46.51 \\
(5) & RandomSampling~+~$40\%$~extra~data & 36.28 & 46.63 \\
\cmidrule{1-4} 
(6) & \modelabbrevname{} ($1^{st}$~round) & 35.29 & 47.13  \\  
(7) & \bf \modelabbrevname{} ($2^{nd}$~round) &\bf  36.41 & \bf 47.38  \\  
\cmidrule[\heavyrulewidth]{1-4}
\end{tabular}
\vspace{-2mm}
\caption{
\small
Test accuracy comparison among variants of \modelabbrevname{} 
on the social relationship classification task. 
}
\label{tab:mainresult2}
\end{center}
\vspace{-5mm}
\end{table}

\begin{figure}[tb]
\centering
\includegraphics[width=0.48\textwidth]
{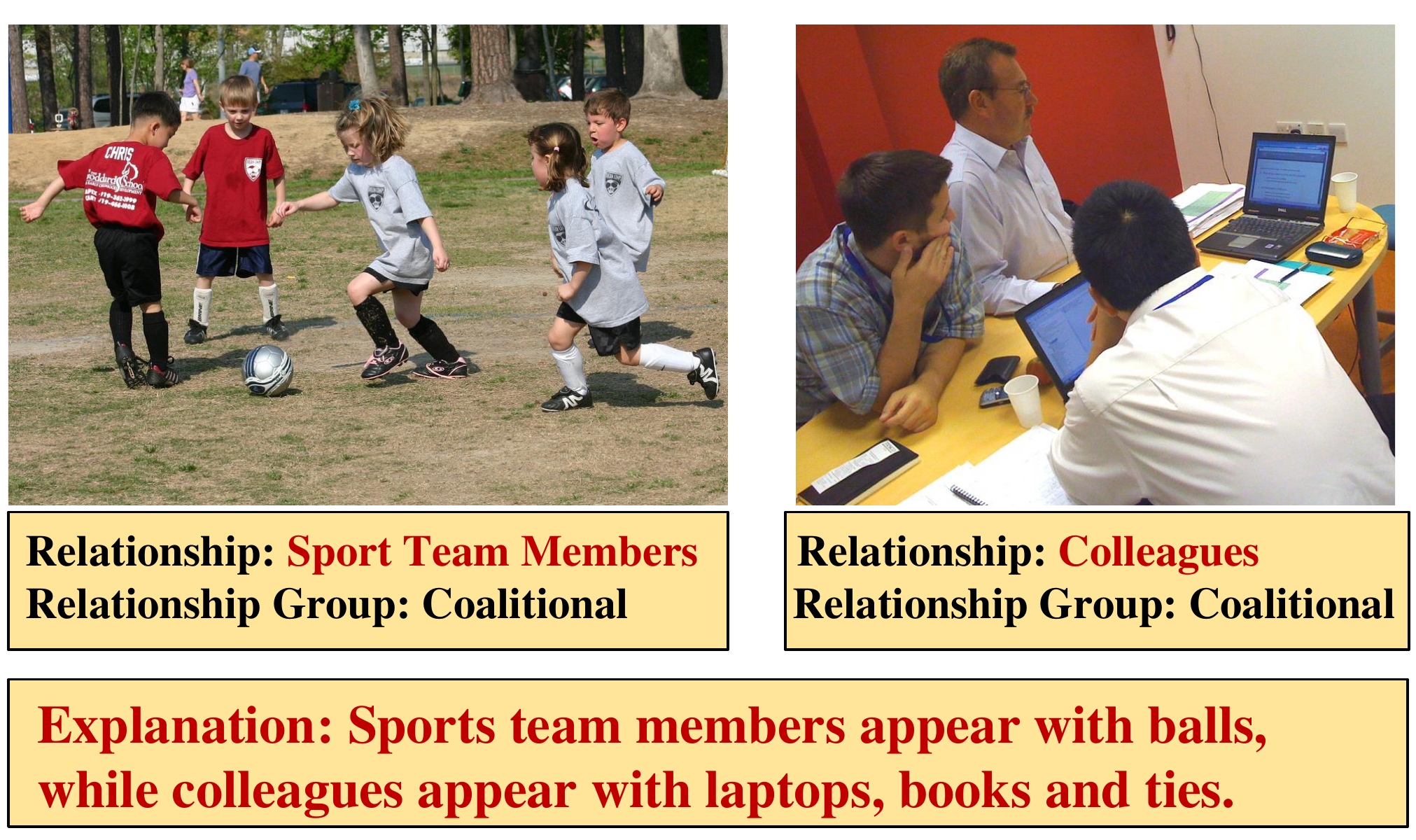} 
\vspace{-6mm}
\caption{
\small 
Examples of social relationship classification. 
Explanations are reconstructed from \citet{DBLP:conf/ijcai/WangCRYCL18}
}
\vspace{-6mm}
\label{fig:dataset2}
\end{figure}

\paragraph{Dataset} 
We also evaluate \modelabbrevname{} on 
the People in Photo Album Relation dataset  ~\cite{DBLP:conf/cvpr/ZhangPTFB15,sun2017domain}. 
An example is shown in Figure~\ref{fig:dataset2}. 
The dataset 
was originally collected from Flickr photo albums and 
involves 5 social domains and 16 social relations. 
We focus on the images that have only two people 
since handling more than two people 
requires task-specific neural architecture. 
The details of dataset pre-processing   
are included in Appendix. 
After pre-processing, 
we obtain $1,679$ training images and $802$ testing images. 
We experiment with a low-resource setting with $15\%$ of the remaining training images (i.e., $264$ images). 
We obtain explanations by 
converting the knowledge graph collected by 
\citet{DBLP:conf/ijcai/WangCRYCL18} 
into a parsed format. 
The semantic segments here are contextual objects like soccer. 
The knowledge graph contains heuristics 
to distinguish social relations by the occurrence of contextual objects (e.g., ``soccer'' for sports v.s. colleagues). 
We use a faster-RCNN-based object detector~\cite{DBLP:journals/pami/RenHG017} trained on the COCO dataset~\cite{DBLP:conf/eccv/LinMBHPRDZ14} 
to localize the semantic segments (contextual objects) during training. 
The object detector is not used during testing. 
We set rounds of expert queries $k=2$ and the query budget $b=4$.

\begin{figure}[tb]
\centering
\includegraphics[width=0.40\textwidth]{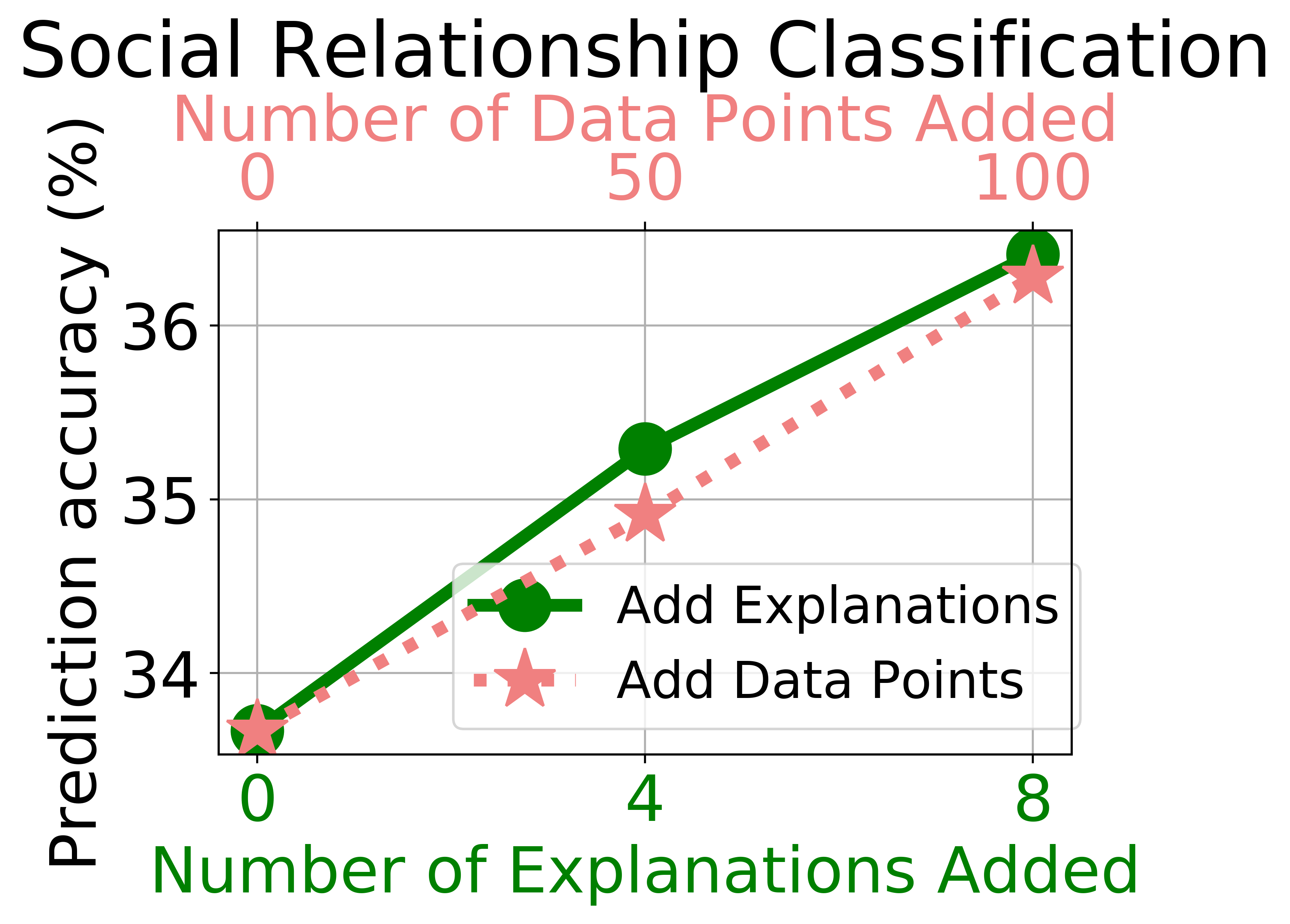}
\caption{
\small
Comparing the performance gain of adding contrastive natural language explanations and adding training data points on social relationship classification. 
}
\label{fig:dataEfficiency2} 
\end{figure}

\paragraph{Results} 
We compare \modelabbrevname{} to its several ablations (Table~\ref{tab:mainresult2}) and evaluate the performance 
on the testing set. 
We report classification accuracy 
on social relationships as well as social domains. 
We observe similar benefits of incorporating explanations to \modelabbrevname{} 
as in the bird species classification task. 
As shown in Table~\ref{tab:mainresult2}, 
the base model with $40\%$ extra training data 
(i.e., $105$ images) 
still slightly 
underperforms \modelabbrevname{} with $8$ explanations 
(RandomSampling~+~$40\%$~extra~data, $36.28\%$ v.s. \modelabbrevname{} ($2^{nd}$~round), $36.41\%$). 
As shown in Figure~\ref{fig:dataEfficiency2}, 
adding $1$ explanation 
leads to similar performance gain 
as adding $13$ labeled training data points. 
Our ablation experiment 
also confirms the importance of 
class-based active learning. 
If we replace class-based active learning with a random selection of class pairs, 
\modelabbrevname{}
learns a bad model structure that leads to reduced performance (\modelabbrevname{} w/ Random Pairs, $22.94\%$). 
The performance drop in domain accuracy is also 
significant. 
We suspect it is because the bad model structure 
confuses the global classifier a lot. 
If the global classifier 
calls a wrong local classifier, 
the local classifier is forced to 
make a prediction on 
such a out-of-distribution data. 
In addition, 
our ablation experiment also verify 
the importance of having knowledge 
beyond having the localization model. 
Substituting the discriminating semantic segments' image patches 
with other semantic segments' patches leads to worse performance 
(\modelabbrevname{} w/ Random Grounding, $27.20\%$). 
One reason is that there are 
many objects in each image. 
Under our low resource setting, 
learning on the image patches of 
random semantic segments may 
make the model to latch on to sample-specific artifacts in the training images, 
which leads to poor generalization.

\section{Conclusion}

We propose an expert-in-the-loop training framework 
\modelabbrevname{} to utilize 
contrastive natural language explanations 
to improve a learning algorithm's data efficiency. 
We extend the concept of active learning to 
class-based active learning for choosing 
the most informative query pair. 
We incorporate the extracted knowledge from expert natural language explanation by changing our algorithm's neural network structure.
Our experiments on two visual recognition tasks 
show that 
incorporating natural language explanations 
is far more data-efficient 
than adding extra training data. 
In the future, 
we plan to examine 
the hierarchical classification architecture's potential for reducing computational runtime.

\section*{Acknowledgments}
We would like to sincerely thank EMNLP 2020 Chairs and Reviewers for their review efforts and helpful feedback. 
We thank Yanbang Wang and Jingjing Tian for their help in early experiments. 
We would also like to extend our gratitude to Zhihua Jin, Xuandong Zhao and Can Liu for their valuable feedback and suggestions.

\bibliography{emnlp2020}
\bibliographystyle{acl_natbib}

\clearpage
\section*{Appendix}

\subsection*{Additional Implementation Details} 
We use Inception v3~\cite{DBLP:conf/cvpr/SzegedyVISW16} as 
our image encoder
$\phi$. 
The global pooling layer $g_{pool}$ is a global average pooling layer. 
The input image size is $(448, 448)$. 
We implement our model in \texttt{PyTorch}. 
We implement the shared attention infrastructure of fine-grained classifiers by 
noting that calculating $QK^T$ 
is equivalent to an efficient $1\times1\times d$ convolution on the activation map 
$\phi(x) \in \mathbb{R}^{H \times W \times d}$, 
with $M$ latent attention queries as $M$ convolutional kernels. 
We use the same hyper-parameters for both datasets. 
We adopt Inception v3~\cite{DBLP:conf/cvpr/SzegedyVISW16} 
as the backbone and choose \texttt{Mix6e} layer as the activation map. 
We tune the hyper-parameters on the unused training images. 
We train the models using Stochastic Gradient Descent
(SGD) with the momentum of $0.9$, weight decay of $1e-5$. 
We decay the learning rate of each parameter group 
by $0.9$ every $2$ epochs using \texttt{torch.optim.lr\_scheduler.StepLR}. 
The global pooling $g$ is a global average pooling layer. 
We set $M$ 
the number of learnable latent attention queries to $6$. 
The total number of parameters of our model is 
$15,114,476$. 
The training time for our approach is less than $20$ minutes 
since our resource constraint setting has a limited amount of training data. 
Unlike previous active learning on data-points, 
our class-based active learning is empirically insensitive 
to the change of random seeds and hyper-parameter (e.g., batch size). 
Therefore, we could collect the explanations in an on-demand manner.

\subsection*{Bird Species Classification Dataset} 
We adopt the random sampling method in ~\cite{DBLP:conf/cvpr/VedantamB0PC17}, 
to make sure that the sampled species are challenging to classify. 
The sampling method is based on birds' biological hierarchy~\cite{DBLP:conf/wacv/BarzD20} from Wikispecies. 
The $25$ randomly sampled bird species are: 
Crested Auklet, 
Least Auklet, 
Parakeet Auklet, 
Tropical Kingbird, 
Gray Kingbird, 
Belted Kingfisher, 
Green Kingfisher, 
Pied Kingfisher, 
Ringed Kingfisher, 
Scarlet Tanager, 
Summer Tanager, 
Brown Thrasher, 
Sage Thrasher, 
California Gull, 
Heermann Gull, 
Ivory Gull, 
Ring billed Gull, 
Black capped Vireo, 
Blue headed Vireo, 
White eyed Vireo, 
Yellow throated Vireo, 
Artic Tern, 
Black Tern, 
Caspian Tern, 
Least Tern.

\subsection*{Saliency Map Visualization} 
We use the techniques in \citet{saliencymap} to visualze the saliency map. 
A saliency map tells us the degree to which each pixel in the image affects the classification score for that image. To compute it, we compute the gradient of the unnormalized score corresponding to the correct class (which is a scalar) with respect to the pixels of the image. If the image has shape $(3, H, W)$ then this gradient will also have shape $(3, H, W)$; for each pixel in the image, this gradient tells us the amount by which the classification score will change if the pixel changes by a small amount. To compute the saliency map, we take the absolute value of this gradient, then take the maximum value over the $3$ input channels; the final saliency map thus has shape $(H, W)$ and all entries are nonnegative.

\subsection*{Social Relationship Classification Dataset} 
PIPA-Relation dataset~\cite{sun2017domain} 
is built on PIPA dataset~\cite{DBLP:conf/cvpr/ZhangPTFB15}. 
We exclude the images with more than two people 
since it requires task-specific neural architecture. 
Since we have annotations of people pairs for each image, 
we could easily identify and remove images with more than two people. 
However, the dataset becomes heavily unbalanced after this step since images of certain relationships tend to have less people. 
To tackle this issue, 
we truncate the classes that have more than $200$ training images left to $200$ training images. 
Similarly, 
we truncate the classes that have more than $100$ testing images left to $100$ testing images. 
We finally get $1679$ training images and $802$ testing images.

\end{document}